\documentclass[10pt]{article}

\usepackage{amsmath,amsfonts}
\usepackage{algorithmic}
\usepackage{algorithm}
\usepackage{array}

\usepackage{tabularx}
\usepackage{makecell}

\usepackage[T1]{fontenc} 
\usepackage[american]{babel}

\usepackage{listings}
\usepackage[colorlinks,urlcolor=blue,linkcolor=blue,citecolor=blue]{hyperref}
\expandafter\def\expandafter\UrlBreaks\expandafter{\UrlBreaks\do\/\do\*\do\-\do\~\do\'\do\"\do\-}
\usepackage{authblk}

\usepackage{textcomp}
\usepackage{stfloats}

\usepackage{verbatim}
\usepackage{graphicx}

\usepackage{xcolor}
\usepackage{tabularx}
\usepackage{makecell}

\usepackage{nameref}

\begin{document}
\title{Visualization of a~multidimensional point cloud as~a~3D swarm of~avatars}

\author{Leszek Luchowski}
\author{Dariusz Pojda}
\affil{Institute of Theoretical and Applied Informatics,\\
Polish Academy of Sciences\\
\texttt{\{lluchowski, dpojda\}@iitis.pl}}

\date{}

\maketitle

\begin{abstract}
This paper proposes an innovative technique for representing multidimensional datasets using icons inspired by Chernoff faces. Our approach combines classical projection techniques with the explicit assignment of selected data dimensions to avatar (facial) features, leveraging the innate human ability to interpret facial traits. We introduce a semantic division of data dimensions into intuitive and technical categories, assigning the former to avatar features and projecting the latter into a four-dimensional (or higher) spatial embedding. The technique is implemented as a plugin for the open-source dpVision visualization platform, enabling users to interactively explore data in the form of a swarm of avatars whose spatial positions and visual features jointly encode various aspects of the dataset. Experimental results with synthetic test data and a 12-dimensional dataset of Portuguese Vinho Verde wines demonstrate that the proposed method enhances interpretability and facilitates the analysis of complex data structures.\end{abstract}

\section{Introduction}

Multidimensional datasets are a very rich resource in many applications, constantly gaining importance as both computer hardware and algorithms are increasingly capable of handling them. One element that is struggling to keep up with this development is the human visual system (and human senses in general), which is naturally built for 3D space and cannot perceive or even imagine a space of higher dimensionality.

There are two ways to cope with the problem: (i) reduce the dimensionality of the data by taking a~3D projection or cross-section (possibly a thick-layer cross-section) of the data; (ii) increase the dimensionality of the display by bestowing the representations of the points with additional characteristics besides their coordinates. Those additional properties can include color, size, shape parameters, and any quantifiable visual peculiarities. Each data point is then represented by a more elaborate token, which in this text will be referred to as an {\it avatar}.

One well-known approach to avatar-based representation is the use of {\it Chernoff faces} \cite{Chernoff1973}, icons resembling a human face, where data values are converted into parameters such as the rounded or oblong shape of the face, hair color, and length, smiling or frowning curvature of lips, etc. Chernoff faces leverage the natural ability of the human observer to identify and interpret facial traits and expressions and have been widely used for intuitive visualization of multivariate data.

However, Chernoff faces present several limitations, especially in the context of complex, high-dimensional, or interactive data analysis. The number of distinct visual features that can be reliably mapped onto a face is inherently limited, constraining the dimensionality that can be visualized. Most implementations are static and two-dimensional, limiting opportunities for dynamic exploration or spatial context. Importantly, the human perception of faces is subject to emotional and subjective biases---for example, certain expressions may be perceived as ''friendly'' or ''untrustworthy'' regardless of the underlying data~\cite{Lardelli}. Furthermore, Chernoff faces typically treat all variables equally without distinguishing between dimensions that are intuitively meaningful to the user and those that are more abstract or technical. It can lead to loss of interpretability or visual clutter, especially as the number of variables increases. To overcome some of these issues, alternative glyphs have been proposed, such as trees~\cite{Lardelli}, which are familiar to humans but evoke less emotion.

\begin{figure}
	\centering
	\includegraphics[width=.4\linewidth]{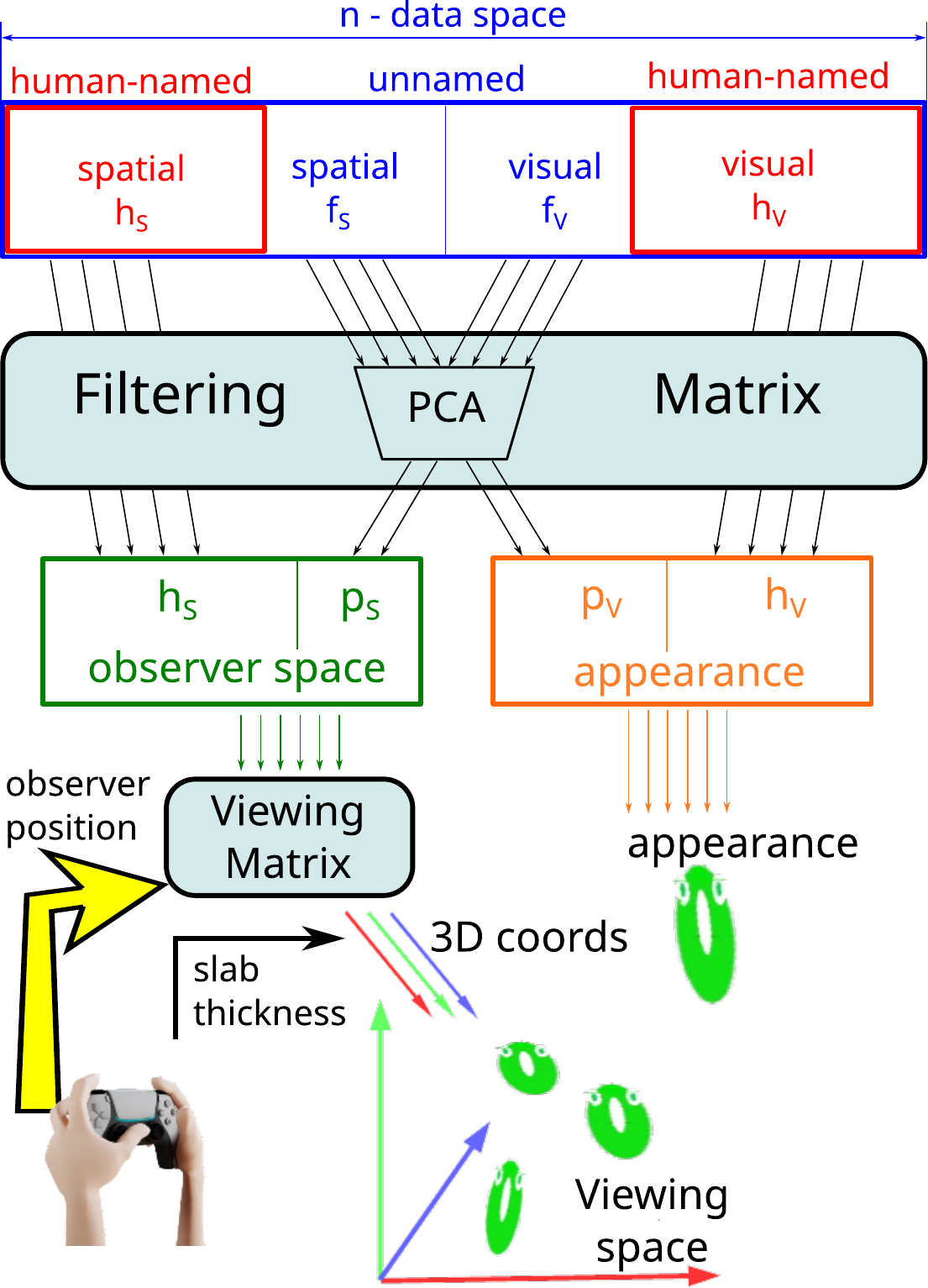}
	\caption{Data dimensions vs display dimensions}
	\label{fig:dimGraph}
\end{figure}

In this work, we propose a novel approach that semantically separates data dimensions into two groups: 
those that are intuitively meaningful to the user (e.g., sensory, categorical, interpretable parameters), and those that are abstract or technical. 
Users may assign meaningful dimensions to either geometric space or avatar visual features.
If fewer than the required number of dimensions are assigned, the system automatically supplements them using anonymous dimensions, which are processed via Principal Component Analysis (PCA) to generate missing spatial or visual features.
This approach is implemented as an interactive plugin for the \textit{dpVision} platform \cite{POJDA2025102093, dpvision}, allowing users to explore a cloud of 3D avatars positioned in a 4D data space and visualized through dynamic projections into 3D space (see Figure~\ref{fig:dimGraph}).
By merging spatial navigation with avatar-based representation, our plugin provides a human-centered and intuitive visualization of complex multidimensional data.

To validate the proposed approach, we use two types of multivariate datasets:
a) two synthetic datasets designed to illustrate interpretability and the effects of variable assignment—one modeling fictional politicians and another simulating a diverse set of soft drinks; and
b) the publicly available dataset containing 12-dimensional data on Portuguese \textit{Vinho Verde} wines~\cite{Cortez09,wine_quality_186}.

The remainder of this paper is organized as follows. 
Section~\ref{sec:related_work} reviews existing methods for multidimensional data visualization and highlights their limitations. 
Section~\ref{sec:methodology} introduces the conceptual framework underlying our approach, including the semantic division of data dimensions, and describes the technical implementation of the visualization system as a plugin for the \textit{dpVision} platform.
Section~\ref{sec:results} presents experiments on both synthetic and real-world datasets and explains the process of interactive exploration of the data cloud.
Section~\ref{sec:discussion} summarizes the main contributions of the work and outlines directions for future research.
Finally, Section~\ref{sec:conclusion} concludes the paper.

\section{Related work}\label{sec:related_work}

The problem of visualizing multidimensional data has inspired a diverse set of techniques, each with its strengths and limitations. Projection-based methods, such as Principal Component Analysis (PCA)~\cite{Jolliffe2011}, t-distributed stochastic neighbor embedding (t-SNE)~\cite{Maaten2008}, and Uniform Manifold Approximation and Projection (UMAP)~\cite{McInnes2018}, are commonly used to reduce high-dimensional datasets to two or three dimensions, facilitating the discovery of patterns, clusters, or outliers.

Coordinate-based visualizations, such as parallel coordinates~\cite{d_ocagne} and radial plots~\cite{Inselberg1985}, allow each variable to be explicitly represented as an axis. Parallel coordinates, in particular, have been widely used in medicine (e.g., ''liver profiles''), and their variants, such as Andrews plots~\cite{Andrews}, provide additional smoothing and compactness. These approaches are effective for moderate numbers of dimensions but can become visually cluttered as dimensionality increases.

Matrix-based and small multiples techniques, including trellis charts~\cite{Cleveland1993}, offer compact overviews of pairwise relationships between variables, enabling users to examine a large number of possible interactions.

Glyph-based approaches, most notably Chernoff faces~\cite{Chernoff1973}, map multiple variables onto visual features of an icon or avatar, leveraging innate human abilities for face and pattern recognition. However, the subjectivity and emotional connotations of human-like glyphs can bias interpretation, and most glyph-based methods are static and limited to two dimensions.

Dynamic and interactive exploration techniques have also been developed, such as the grand tour and guided tour~\cite{Cook05} and the slice tour~\cite{Cook2020SliceTour}. These methods present users with sequences of projections or thin slices through the data space, sometimes augmented with glyphs for detailed inspection (e.g., the \texttt{tourr} package for R). While powerful for uncovering high-dimensional structures, these approaches typically treat all variables as equivalent, making them challenging for non-experts to interpret and understand.

More recently, cognitively grounded approaches have been explored. Pflughoeft et al.~\cite{PFLUGHOEFT2024103911} introduced ''data avatars''---metaphorical glyphs designed to visually encode multidimensional data in ways that align with intuitive human perception. Their work, based on dual-process cognitive theory, demonstrated that such designs could improve interpretation; however, they did not address interactive navigation or explicit semantic division of variables.

Compared to the above methods, our approach introduces two key innovations. First, we enable a semantic split of the dataset into two groups: dimensions that are intuitively meaningful to users (such as sensory or quality ratings) and those that are more technical or abstract (such as chemical or process parameters). The intuitive features are mapped to avatar visual traits (inspired by Chernoff faces), while the technical features are projected into spatial coordinates, optionally using dimensionality reduction (e.g., PCA). Second, we integrate these glyphs into an interactive 3D environment, allowing users to explore a ''swarm'' of avatars, dynamically rotate the viewpoint, and apply slab-based filtering to reveal local structure.

Our slab filtering is conceptually related to the slice tour~\cite{Cook2020SliceTour}, as both select points near a hyperplane for visualization. However, while the slice tour typically focuses on very thin slices (showing only points lying almost exactly on a hyperplane), our approach uses a slab of configurable thickness, retaining all points within a certain distance from the projection plane. This design supports the practical exploration of moderately sized datasets, reducing visual clutter while still revealing structural patterns. To our knowledge, this is the first system to combine avatar-based glyphs with interactive 3D navigation, explicit semantic variable assignment, and flexible local filtering in this manner.

\section{Methodology}\label{sec:methodology}

Throughout this paper, we use consistent notation to describe data dimensions, transformation matrices, and visualization parameters. The key symbols and abbreviations are summarized in Table~\ref{tab:symbols} and are referenced throughout the description of the projection procedure and implementation details.

\begin{table}[t!]
\small
\renewcommand{\arraystretch}{1.2}
\centering
\caption{Symbols and notation used throughout the paper.}
\label{tab:symbols}
\begin{tabularx}{\linewidth}{>{\centering\arraybackslash}m{0.7cm} X >{\centering\arraybackslash}m{2.7cm}}
\hline
\textbf{Symbol} & \textbf{Meaning} & \textbf{Constraints} \\
\hline
\multicolumn{3}{c}{\textbf{Input data parameters}} \\
\hline
$N$ & Total number of data points & ~ \\
$n$ & Total number of data dimensions & ~ \\
$X$ & Original data matrix & $X \in \mathbb{R}^{n \times N}$ \\
\hline
\multicolumn{3}{c}{\textbf{Implementation-dependent mapping}} \\
\hline
$k$ & Number of navigable display dimensions & \makecell{$k \geq 3$\\(typically 4)} \\
$m$ & Number of available avatar visual features & \makecell{$m \leq n - k$\\(currently 10)}\\
\hline
\multicolumn{3}{c}{\textbf{User's mapping}} \\
\hline
$h_S$ & Number of human-meaningful dimensions mapped to space & $h_S \leq k$ \\
$h_V$ & Number of human-meaningful dimensions mapped to avatars & $h_V \leq m$ \\
$h_A$ & Number of anonymous dimensions & $h_A \leq n - (h_S+h_V)$ \\
\hline
\multicolumn{3}{c}{\textbf{Projection matrices}} \\
\hline
$F$ & Filtering matrix & $F \in \mathbb{R}^{(k + m) \times n}$ \\
$V$ & \makecell[l]{View matrix: for spatial rotation in $k$-dimensional\\
space ($k\times k$), or full affine transformation with\\
translation in homogeneous coordinates\\ ($(k+1)\times(k+1)$)} & \makecell{$V \in \mathbb{R}^{k \times k}$ or\\ $V \in \mathbb{R}^{(k+1) \times (k+1)}$} \\
\hline
\multicolumn{3}{c}{\textbf{Derived datasets}} \\
\hline
$X_{\text{spatial}}$ & Spatial projection of $X$ & $X_{\text{spatial}} \in \mathbb{R}^{k \times N}$ \\
$X_{\text{visual}}$ & Avatar feature mapping of $X$ & $X_{\text{visual}} \in \mathbb{R}^{m \times N}$ \\
$X_{\text{view}}$ & Spatial data after applying $V$ & $X_{\text{view}} \in \mathbb{R}^{k \times N}$ \\
\hline
\end{tabularx}
\end{table}

\subsection{Paradigm and semantic assignment of variables}\label{sec:paradigm}

In practical applications, multidimensional datasets often contain both variables that are intuitively meaningful to users (such as sensory ratings and categorical or interpretable parameters) and features that are more technical or abstract (such as chemical coefficients or structural process data). To maximize interpretability, our approach introduces a \textit{semantic division} of data dimensions into two groups:
\begin{itemize}
    \item \textbf{Intuitive dimensions}---readily interpreted by users---are mapped onto visual characteristics of avatars (e.g., face shape, expressions, color).
    \item \textbf{Technical dimensions}---more abstract or difficult to interpret---are projected into spatial coordinates in the observation space, possibly after dimensionality reduction (e.g., PCA).
\end{itemize}

The assignment of variables to these groups should be guided by domain knowledge and the analysis's objectives. Features that are critical for individual profiling or human interpretation should be mapped to avatar features, while those essential for discovering structures or clusters may be included in spatial coordinates. Redundancy should be avoided (e.g., highly correlated features mapped to multiple avatar traits), and important variables should not be left out of both avatar and spatial assignment, as this could diminish interpretability.

\begin{figure}[t!]
    \centering
    \includegraphics[width=0.4\linewidth]{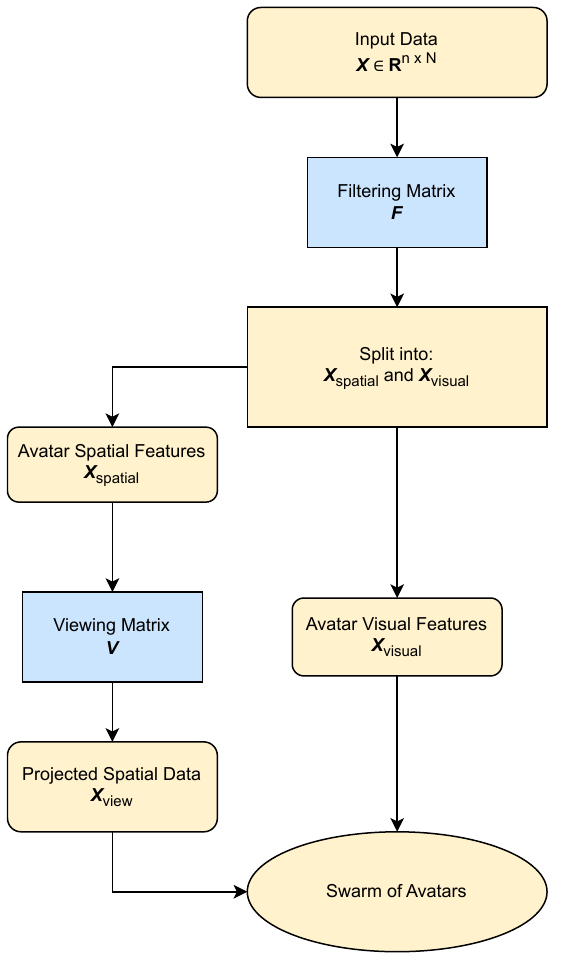}
    \caption{Conceptual diagram illustrating the separation of multidimensional data into spatial coordinates and avatar visual features through filtering and viewing transformations.}
    \label{fig:paradigm_diagram}
\end{figure}

Let the original dataset be represented by a matrix $X \in \mathbb{R}^{n \times N}$, where $n$ is the number of original dimensions and $N$ is the number of data points.

To implement the semantic assignment of variables, we construct a filtering matrix $F \in \mathbb{R}^{(k + m) \times n}$ that separates and processes the dimensions, where $k$ is the number of spatial dimensions and $m$ is the number of avatar visual features. The filtered data are given by Eq.~\eqref{eq:filtered}:
\begin{equation}
    X_{\text{filtered}} = F \cdot X.
    \label{eq:filtered}
\end{equation}
Each row of $F$ defines a new output feature (spatial or visual) as a linear combination of the original dimensions of $X$.

Spatial coordinates are obtained by selecting the appropriate subset $F_{\text{spatial}} \in \mathbb{R}^{k \times n}$, as shown in Eq.~\eqref{eq:spatial}:
\begin{equation}
    X_{\text{spatial}} = F_{\text{spatial}} \cdot X,
    \label{eq:spatial}
\end{equation}
while visual parameters are derived from $F_{\text{visual}} \in \mathbb{R}^{m \times n}$, according to Eq.~\eqref{eq:visual}:
\begin{equation}
    X_{\text{visual}} = F_{\text{visual}} \cdot X.
    \label{eq:visual}
\end{equation}

Finally, the spatial data are further transformed for display using a view matrix $V$, as in Eq.~\eqref{eq:view}:
\begin{equation}
    X_{\text{view}} = V \cdot X_{\text{spatial}}
    \label{eq:view}
\end{equation}
where $V$ is typically a $k \times k$ rotation matrix. In practical applications, particularly for interactive 3D or 4D exploration, we use an affine $(k+1)\times(k+1)$ matrix acting on homogeneous coordinates, combining both rotation and translation (see Implementation).

This dual mapping---avatars for intuitive features, spatial positioning for technical features---provides a flexible and human-centered way to visualize high-dimensional data (see Figure~\ref{fig:paradigm_diagram}).

\textbf{Guidelines and limitations.} The semantic assignment should be made carefully, considering both user needs and analytical goals. Important features should not be omitted or relegated solely to PCA-based projection, as their interpretability may be lost (principal components represent blends of original variables). The division between intuitive and technical features is context-dependent and may vary by task or user group. We recommend iterative testing of assignments, consultation with domain experts, and documenting mapping choices. When PCA is used to process unassigned (anonymous) variables, users should inspect principal component loadings to understand which original features contribute most strongly to the resulting spatial dimensions. Although our current tool does not include interactive visualization of these loadings, such analysis can be performed using standard statistical software.

This framework strikes a balance between the interpretability of human-centered visualization and the structural insight provided by dimensionality reduction, remaining transparent about the trade-offs involved.

\subsection{Implementation}\label{sec:implementation}

Based on the conceptual framework presented in the previous section, the practical implementation of the proposed visualization method involves constructing appropriate transformation matrices, performing dimensionality reduction where necessary, and assigning selected data dimensions to spatial coordinates or avatar visual features.

The implementation uses the Eigen library (C++), which allows efficient operations on matrices and vectors.

The input data are stored in a matrix of size \(n \times N\):
\begin{lstlisting}[language=C++]
Eigen::MatrixXd X(n, N);
\end{lstlisting}
Each column corresponds to a single point in the data space \(\mathbb{R}^n\), and each row represents one of the \(n\) dimensions in that space.

\subsubsection{Assigning dimensions to categories}

\textbf{Note:} To avoid confusion, we refer to \emph{dimensions} when describing input data and to \emph{features} when referring to the visualized output.

\begin{figure}[b!]
	\centering
	\includegraphics[width=.4\linewidth]{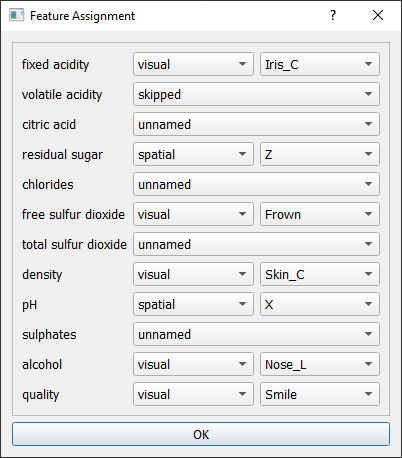}
	\caption{Dialog window for assigning input dimensions to feature categories: \textit{spatial}, \textit{visual}, \textit{unnamed}, or \textit{skipped}. For spatial and visual categories, a specific output feature label can be selected.}

	\label{fig:dialog}
\end{figure}

Input dimensions can be assigned to one of three feature categories: \textit{spatial}, \textit{visual}, or \textit{anonymous}. Additionally, any dimension may be marked as \textit{skipped} and excluded from further processing.

This assignment is performed through a dedicated dialog window (Figure~\ref{fig:dialog}), which enables the user to associate input dimensions with specific feature categories and, when applicable, select predefined visual or spatial feature labels.

The complete set of available labels for spatial and visual features is presented in Table~\ref{tab:features}.

The resulting configuration is stored in index tables. It is not required to assign all available features---any missing spatial or visual values are automatically interpolated during subsequent processing.

\begin{table}[ht]
\small
\renewcommand{\arraystretch}{1.2}
\centering
\caption{List of features used in the implementation}
\label{tab:features}
\begin{tabularx}{\linewidth}{l l X}
\hline
\textbf{Category} & \textbf{Feature} & \textbf{Description} \\
\hline
\multicolumn{3}{l}{\textbf{User-defined spatial features}} \\
\hline
spatial & \texttt{X} & X coordinate in 4D space \\
spatial & \texttt{Y} & Y coordinate in 4D space \\
spatial & \texttt{Z} & Z coordinate in 4D space \\
spatial & \texttt{T} & T coordinate in 4D space \\
\hline
\multicolumn{3}{l}{\textbf{User-defined avatar facial attributes}} \\
\hline
visual & \texttt{Skin\_C} & avatar’s skin color \\
visual & \texttt{Hair\_C} & avatar’s hair color \\
visual & \texttt{Eye\_S} & distance between the avatar’s eyes \\
visual & \texttt{Nose\_L} & avatar’s nose length \\
visual & \texttt{Mouth\_W} & avatar’s mouth width \\
visual & \texttt{Smile} & degree of curvature of the avatar’s smile \\
visual & \texttt{Frown} & level of frowning expression \\
visual & \texttt{Hair\_L} & avatar’s hair length \\
visual & \texttt{Face\_Elong} & avatar’s face proportions \\
visual & \texttt{Iris\_C} & avatar’s iris color \\
\hline
\multicolumn{3}{l}{\textbf{features not defined by the user}} \\
\hline
anonymous & ~ & used for principal component analysis (PCA) \\
\hline
\multicolumn{3}{l}{\textbf{Features selected by the user to skip}} \\
\hline
skipped & ~ & excluded from further processing \\
\hline
\end{tabularx}
\end{table}

\subsubsection{Filtering matrix}

The first step, corresponding to Eq.~\eqref{eq:filtered}, is to create the filtering matrix, which in the implementation has the following form:

\begin{lstlisting}[language=C++]
Eigen::MatrixXd F(k + m, n);
F.setZero();
\end{lstlisting}

To simplify later operations, we assume that the first \(k\) rows of this matrix correspond to spatial features (in our implementation: X, Y, Z, T), and the following \(m\) rows correspond to visual features (e.g., avatar appearance details). This separation is formally described by Eqs.~\eqref{eq:spatial} and~\eqref{eq:visual}. The matrix \(F\) is initially filled with zeros and then completed based on the feature assignments:

\begin{itemize}
  \item First, for each of the \(h_S\) spatial features and \(h_V\) visual features selected by the user, one is set at the intersection of the row for that feature and the column of the original input dimension.
  
  \item Then, if there are still unused rows in \(F\), and the user has selected anonymous features (\(h_A > 0\)), a PCA is performed on the input dimensions assigned to those anonymous features. The \((k + m) - (h_S + h_V)\) strongest components are selected, and their coefficient vectors are inserted into the remaining empty rows of \(F\).
\end{itemize}

The transformed data is calculated as the product of the filtering matrix and the input matrix, followed by normalization. The procedure for creating the filtering matrix and transforming the input data is shown in Algorithm~\ref{alg:projection}.

\begin{algorithm}
\caption{Implementation of Data Projection}
\label{alg:projection}
\begin{algorithmic}[1]
\REQUIRE $X$ (all features), $Asgn$ (assignment of features to groups)
\ENSURE $X_{\text{filtered}}$ (projected data)

\STATE Initialize $F$ (zero matrix)

\FOR{each feature index $j$}
    \IF{assignment of feature $j$ is \textbf{ignored}}
        \STATE Skip
    \ELSIF{assignment of feature $j$ is \textbf{spatial}}
        \STATE Mark corresponding spatial row in $F$
    \ELSIF{assignment of feature $j$ is \textbf{visual}}
        \STATE Mark corresponding visual row in $F$
    \ELSIF{assignment of feature $j$ is \textbf{anonymous}}
        \STATE Save index of feature $j$ for PCA
    \ENDIF
\ENDFOR

\IF{number of mapped features $(h_S + h_V) < (k + m)$}
    \STATE Perform SVD on anonymous features
    \STATE Select $(k + m) - (h_S + h_V)$ top components
    \STATE Insert principal directions into empty rows in $F$
\ENDIF

\STATE $X_{\text{filtered}} = F \times X$

\FOR{each row in $X_{\text{filtered}}$}
    \STATE Center and scale (standardize)
    \IF{row corresponds to spatial dimension}
        \STATE Shift by $+0.5$
    \ENDIF
\ENDFOR

\RETURN $X_{\text{filtered}}$

\end{algorithmic}
\end{algorithm}

In this way, we obtain a table of \(N\) points, each with \((k + m)\) dimensions. These dimensions can be divided into two categories:

\begin{itemize}
    \item The first \(k\) rows form the matrix \texttt{X\_spatial}, which contains spatial features.
    
    \item The following \(m\) rows form the matrix \texttt{X\_visual}, which contains avatar visual features.
\end{itemize}

\begin{lstlisting}[language=C++]
Eigen::MatrixXd X_filtered(k + m, N);
X_filtered = F * X;

Eigen::MatrixXd X_spatial = X_filtered.topRows(k);
Eigen::MatrixXd X_visual = X_filtered.bottomRows(m);
\end{lstlisting}

The matrix \(\mathbf{X}_{\text{spatial}}\) is now a ready-to-use \(k\)-dimensional dataset for further processing.

\subsubsection{View matrix}

The view matrix $V$ is implemented as an affine $(k+1)\times(k+1)$ matrix, combining spatial rotation (the $k \times k$ upper-left block) and translation (the final column) for interactive navigation in homogeneous coordinates. This approach enables efficient and flexible real-time manipulation of the data cloud in 3D or 4D space.

In the current plugin implementation, all 4D rotations are carried out in coordinate planes such as XY, XZ, XT, YZ, YT, or ZT. These correspond to classic SO(2) rotations within two-dimensional subspaces. However, the implementation can be extended to support full SO(4) rotations, offering a more expressive and flexible navigation model.

The remaining part of the view matrix stores a translation vector and a homogeneous row to enable affine transformations.

In code, the matrix can be built as follows:

\begin{lstlisting}[language=C++]
// Rotation only (k x k)
Eigen::MatrixXd R = generateRotationMatrix(k, axis1, axis2, angle);

// Create full (k+1 x k+1) view matrix with translation
Eigen::MatrixXd V = Eigen::MatrixXd::Identity(k + 1, k + 1);
V.topLeftCorner(k, k) = R;
V.topRightCorner(k, 1) = translation_vector; // optional
\end{lstlisting}

To apply the transformation, the spatial data must be converted into homogeneous coordinates by adding a row of ones:

\begin{lstlisting}[language=C++]
Eigen::MatrixXd X_hom(k + 1, N);
X_hom.topRows(k) = X_spatial;
X_hom.row(k) = Eigen::RowVectorXd::Ones(N);

Eigen::MatrixXd X_view = V * X_hom;
\end{lstlisting}

This corresponds directly to the transformation defined in Eq.~\eqref{eq:view}, where the spatial coordinates are mapped into the observation space for visualization.  
This procedure enables smooth 4D navigation, including both rotation and panning of the data cloud in the observation space.

\subsubsection{Slab-based Filtering}

To limit the number of visible data, a slab-based filtering mechanism is employed. The goal is to retain only those data points that lie close to the viewing space---within a given distance \texttt{slab\_thickness} from a hyperplane defined by the current rotation.

The filtering uses a normal vector to the viewing space to compute signed distances from each point. Depending on whether the filtering is applied before or after transforming the data by the view matrix, different normal vectors are used:
\begin{itemize}
    \item if filtering is done in the original spatial coordinates (\texttt{X\_spatial}), the normal vector can be taken as the last row of the rotation matrix, which defines the orientation of the view;
    
    \item if filtering is done after applying the rotation (\texttt{X\_view}), the normal vector can be defined in canonical coordinates as \([0, 0, 0, 1]\), which corresponds to the last axis.
\end{itemize}

Each point is projected onto the normal vector, and only those within the allowed threshold are retained.

\begin{lstlisting}[language=C++]
Eigen::VectorXd slab_values = normal_vector.transpose() * X_view;
auto mask = slab_values.cwiseAbs().array() < slab_threshold;
Eigen::MatrixXd X_visible = skip_masked(X_view, mask);
\end{lstlisting}

\begin{figure}
	\centering
	\includegraphics[width=.4\linewidth]{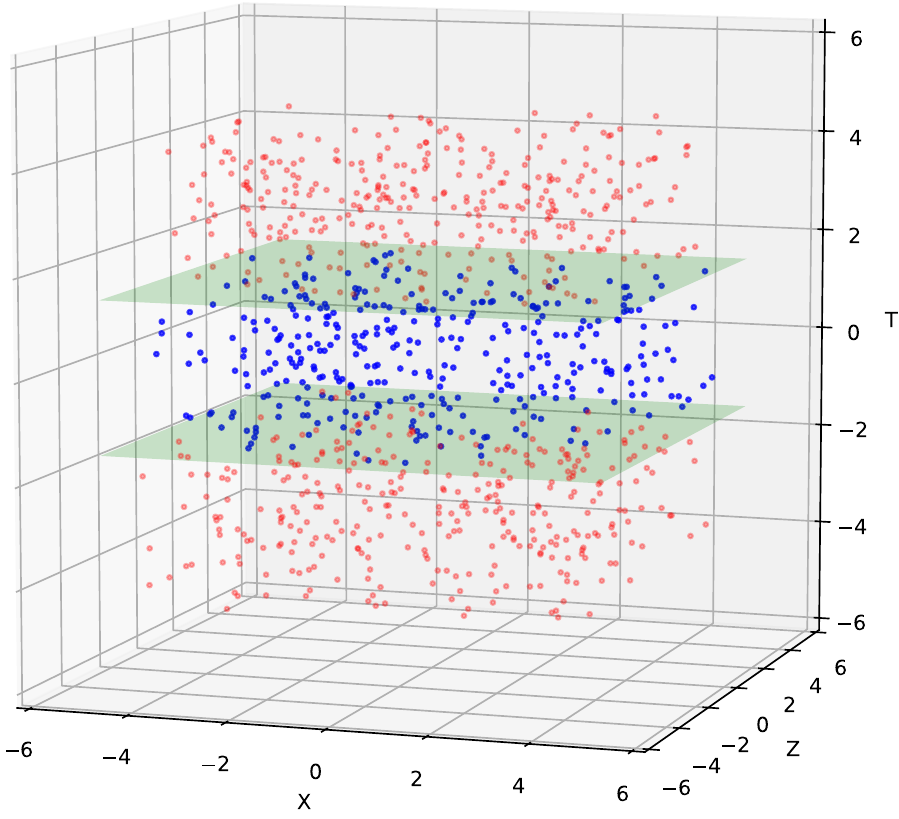}
	\caption{A cloud of points intersected by a slab, using a threshold of 1.5. Red points that lie outside the slab are discarded.}
	\label{fig:slab}
\end{figure}

Points that satisfy the slab condition---meaning they are sufficiently close to the viewing hyperplane---are passed to the next stage of the pipeline. This filtering effectively creates a \textit{thick slice} through the multidimensional space (Figure~\ref{fig:slab}).

\subsubsection{3D projection}

The four-dimensional coordinates are then projected into 3D space using a standard perspective projection. Avatars are generated as three-dimensional objects based on visual features stored in the \(\mathbf{X}_{\text{visual}}\) matrix and rendered using the tools provided by the \textit{dpVision} platform. To ensure interactive performance with large datasets, the avatars are deliberately simplified in terms of geometry, allowing a large number of them to be rendered and manipulated in real time.

\section{Results}\label{sec:results}

In this section, we present experimental results on both synthetic and real-world datasets to illustrate the interpretability and practical utility of the proposed avatar-based multidimensional visualization method. The analyses focus on (i) how intuitive and technical variables are assigned and visualized, (ii) what insights spatial avatars provide compared to classical approaches, and (iii) the impact of variable assignment choices.

\subsection{Small-scale illustration: Synthetic politicians dataset (N=12)}\label{sec:politicians}

This illustrative example utilizes a manually constructed synthetic dataset of 12 fictional politicians, each described by nine features that reflect both voter perception and performance metrics. Intuitive features (number of promises, fulfillment rate, public sympathy, general popularity) were mapped to avatar visual traits (e.g., smile, nose length, skin color). In contrast, technical features (sympathy, economic views, social views) were assigned to spatial coordinates (X, Y, Z axes). The remaining variables (media activity, voting effectiveness, age) were left anonymous for PCA. The variable \texttt{group\_numeric} encodes four archetypes (populists, conservatives, liberals, technocrats), used only for color coding of avatar hair; each sample is labeled by name.

\begin{figure}[h!]
    \centering
    \includegraphics[width=.75\linewidth]{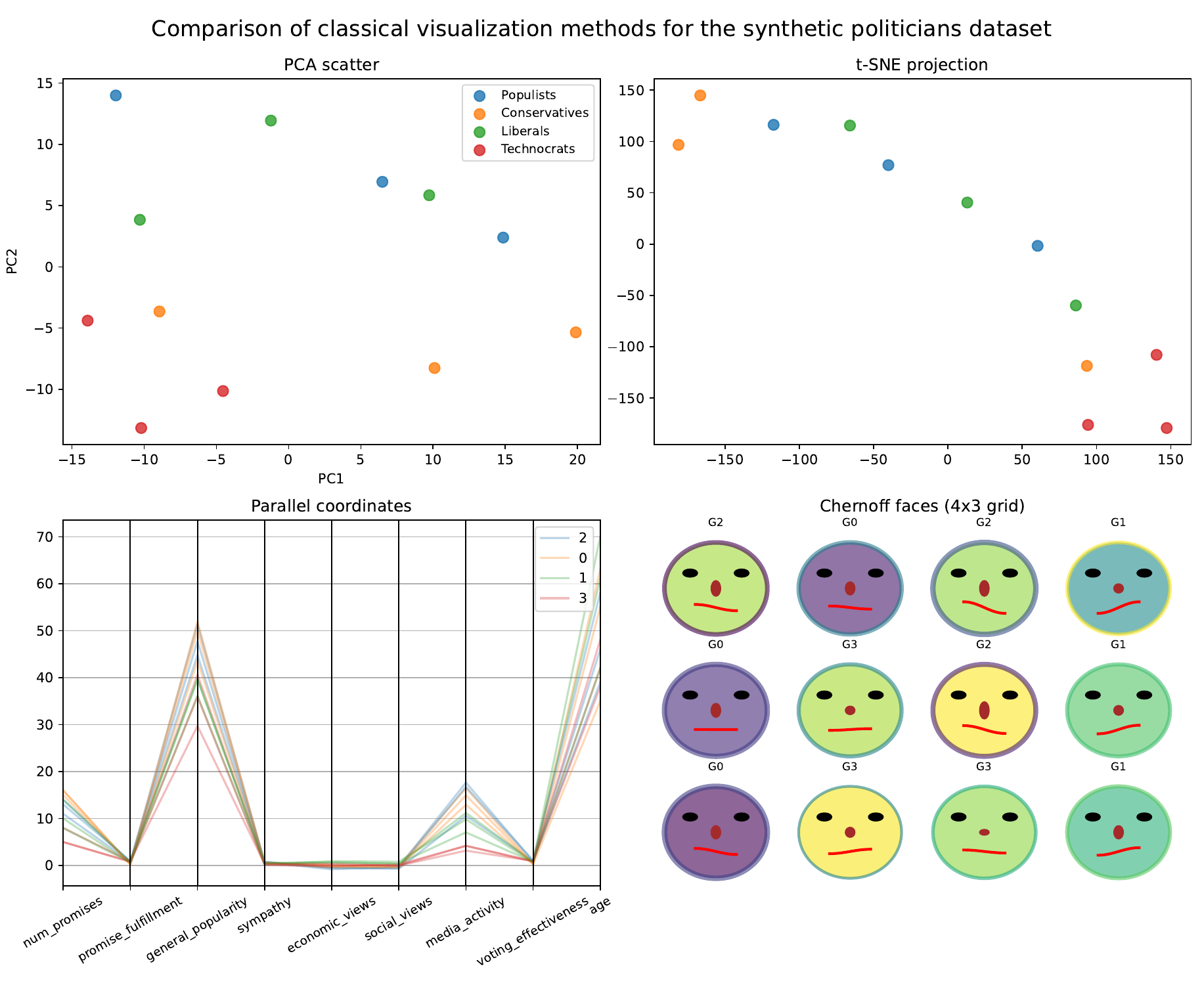}
    \caption{
        Comparison of classical visualization methods for the synthetic politicians dataset.  
        \textbf{Top left:} PCA scatter plot (first two components);  
        \textbf{Top right:} t-SNE projection;  
        \textbf{Bottom left:} parallel coordinates plot;  
        \textbf{Bottom right:} Chernoff faces grid (4$\times$3) for all twelve politicians, colored and labeled by group.
        Each approach reveals different aspects of the data: scatter plots show grouping structure, parallel coordinates display multivariate profiles, and Chernoff faces summarize intuitive features---but none allows simultaneous interpretation of both structure and profile as directly as spatial avatars (see Fig.~\ref{fig:politicians_swarm}).
    }
    \label{fig:politicians_classical}
\end{figure}

To provide a fair comparison with established visualization techniques, Figure~\ref{fig:politicians_classical} presents the same dataset using four classical methods: PCA scatter plot, t-SNE projection, parallel coordinates, and a 4$\times$3 grid of Chernoff faces. While PCA and t-SNE reveal the grouping structure, and parallel coordinates allow for the comparison of individual profiles, none of these methods offers the same level of simultaneous, intuitive insight into both position and feature profiles as spatial avatars. In particular, Chernoff faces communicate intuitive attributes but lack spatial context; conversely, scatter plots provide structure but obscure the source of differences between cases.

For the twelve synthetic politicians, classical methods highlight some group separation (notably in PCA and t-SNE), but interpreting individual differences remains challenging. The Chernoff faces grid allows for visual comparison of intuitive features, yet it does not convey the geometric relationships between politicians or facilitate the detection of outliers in attribute space. This contrast underscores the unique advantage of spatial avatars in combining the interpretability of individual profiles with structural awareness.

\paragraph{Variable assignment:}
\textbf{Avatar features:} number of promises (skin color), promise fulfillment (nose length), general popularity (smile), group affiliation (hair color).
\textbf{Spatial coordinates:} sympathy (X), economic views (Y), social views (Z).
\textbf{Anonymous/PCA:} media activity, voting effectiveness, age.

\begin{figure}[h!]
    \centering
    \includegraphics[width=.7\linewidth]{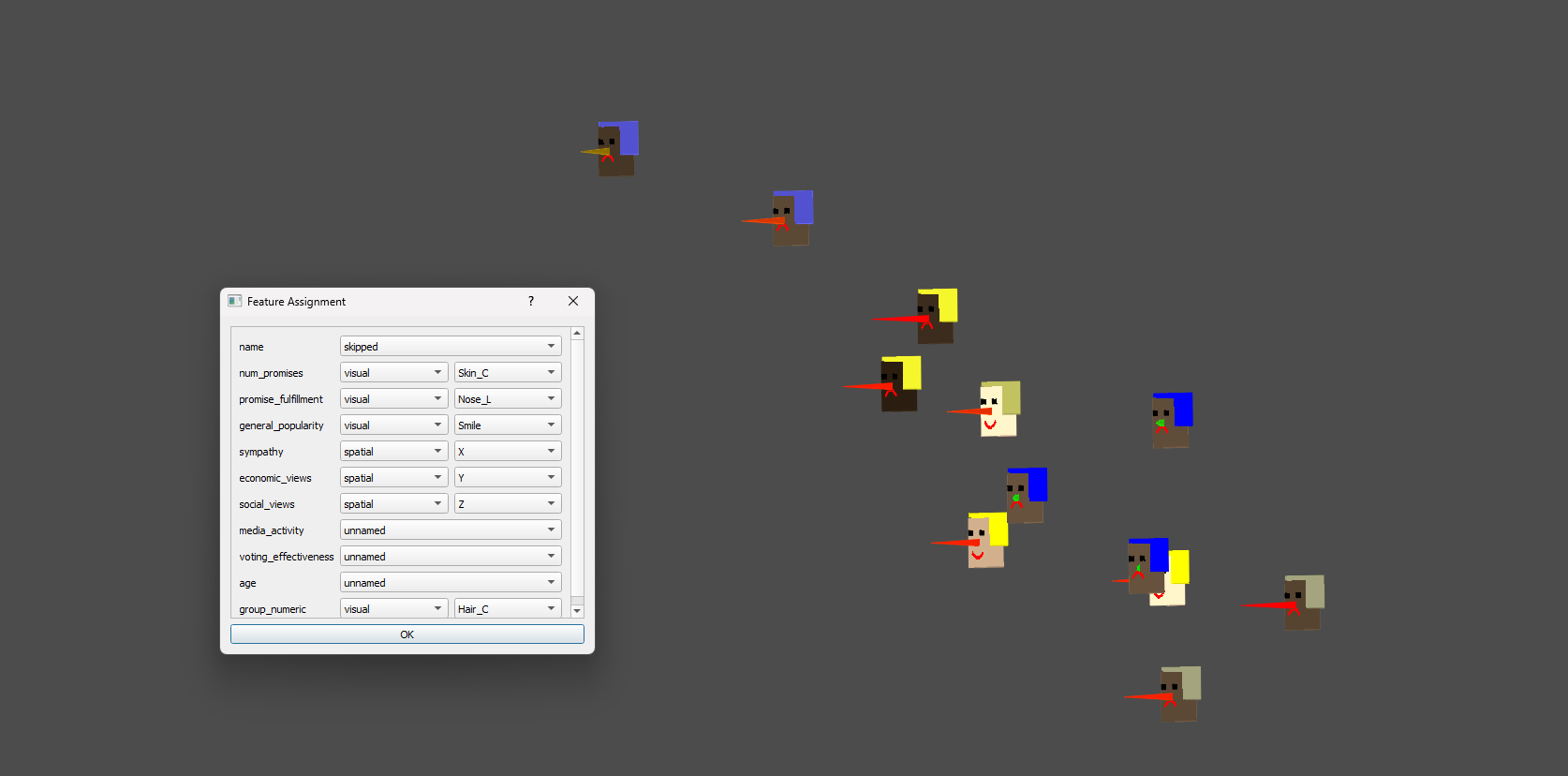}
    \caption{
        Spatial avatar visualization of the synthetic politicians dataset.
        Avatar features encode intuitive traits (skin color: number of promises; nose length: promise fulfillment; smile: general popularity; hair color: group).  
        Spatial coordinates reflect sympathy (X), economic views (Y), and social views (Z).  
        Names and group colors aid interpretation.  
        This visualization enables direct comparison of political archetypes, identification of outliers, and exploration of individual profiles.
    }
    \label{fig:politicians_swarm}
\end{figure}

\paragraph{Interpretation and interactive exploration:}
The spatial avatar visualization enables immediate and intuitive comparison of both the position and the profile of each fictional politician. Clusters and polarization become visible in space, while characteristic features (such as promise fulfillment or popularity) are instantly ''readable'' from the avatar's face. Group affiliation is recognized by hair color.  
Importantly, the tool supports interactive exploration, allowing users to rotate, zoom, and navigate the scene to examine clusters, compare individuals, and discover patterns from multiple perspectives. This interactivity facilitates flexible hypothesis testing and nuanced understanding---features not available in static PCA scatter plots.

\begin{figure}[h!]
    \centering
        \includegraphics[width=.75\linewidth]{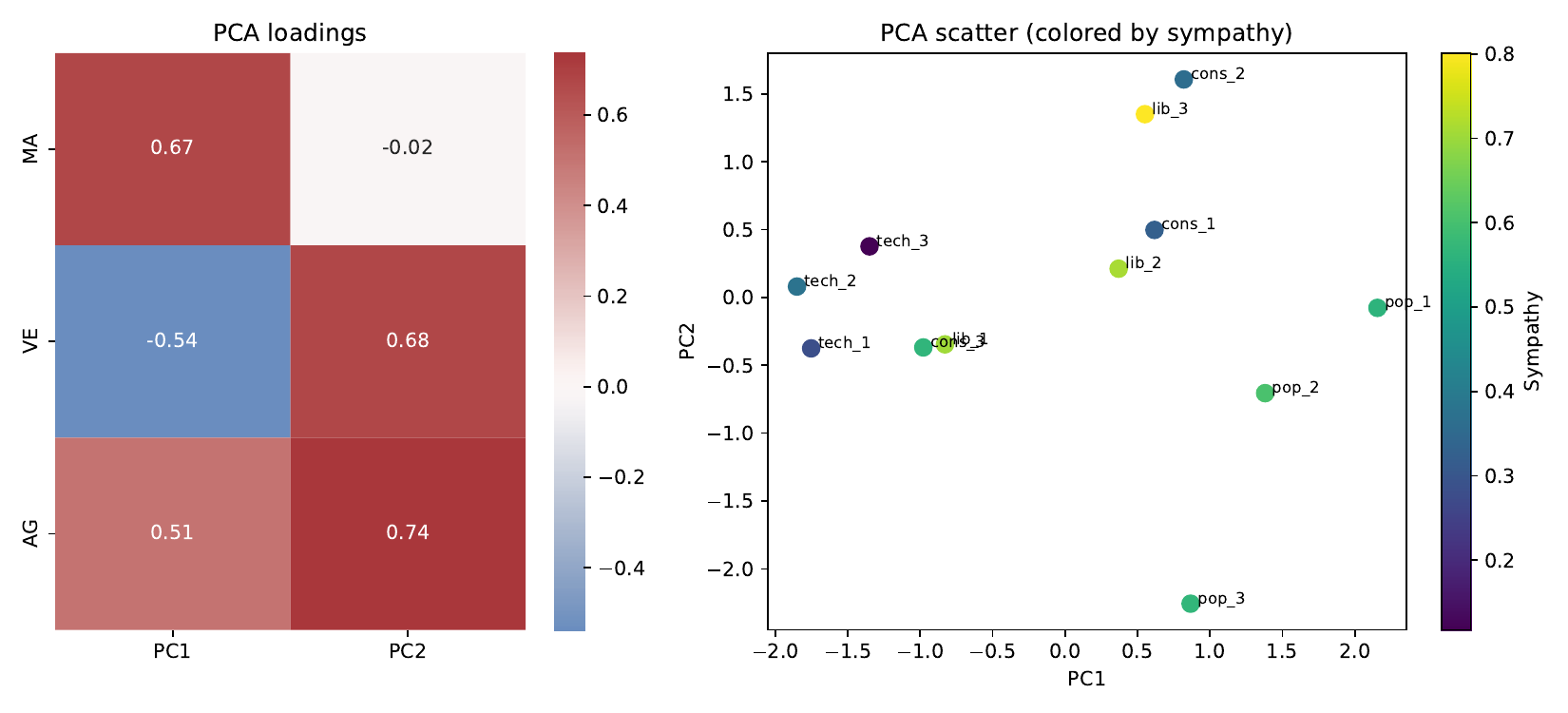}\\
        \includegraphics[width=.75\linewidth]{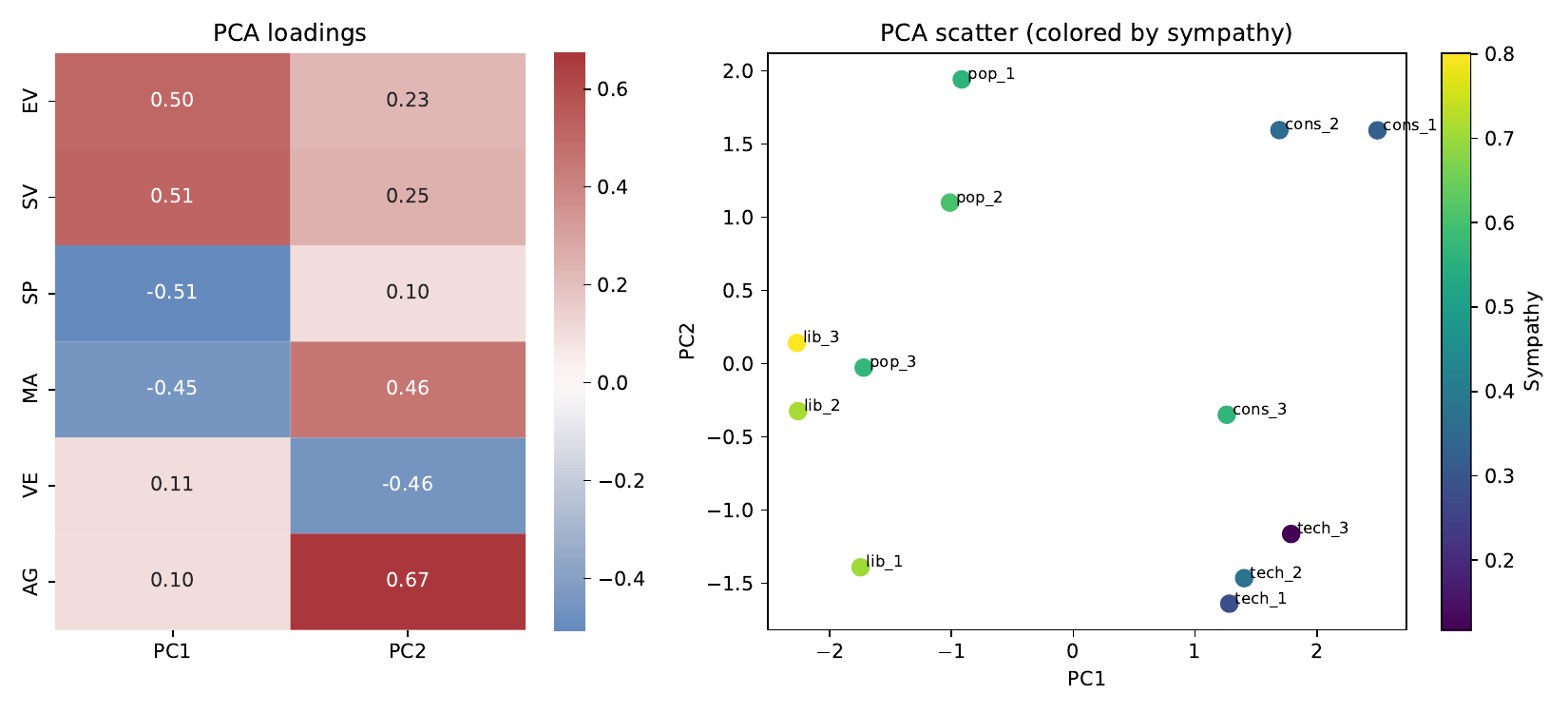}
    \caption{
        Effect of variable assignment on principal component analysis for the synthetic politicians dataset.
        \textbf{(top)} PCA on anonymous features only.
        \textbf{(bottom)} PCA on all features assigned to spatial coordinates or left anonymous (all except avatar features).
        Variable abbreviations: MA---media activity, VE---voting effectiveness, AG---age, SY---sympathy, EV---economic views, SV---social views, SP---sympathy.
    }
    \label{fig:politicians_pca_analysis}
\end{figure}

\paragraph{PCA loadings and spatial interpretation:}
Principal Component Analysis was performed under two scenarios: (top) only on anonymous features (not mapped to avatars or spatial coordinates); (bottom) on all features assigned to spatial coordinates or left anonymous (all except avatar features).  
Figure~\ref{fig:politicians_pca_analysis} presents heatmaps of PCA loadings and the corresponding scatter plots.

The PCA scatter plot in Figure~\ref{fig:politicians_classical} (top left) provides a classical projection of the entire feature set for all politicians, serving as a baseline for comparison.
In contrast, Figures~\ref{fig:politicians_pca_analysis}a and~b show PCA performed on, respectively, (a) only anonymous features not mapped to avatars or spatial coordinates and (b) all features mapped to spatial coordinates or left anonymous. These alternative PCA projections illustrate how the interpretability of the spatial arrangement depends critically on which features are included in the dimensionality reduction.

In both PCA scenarios, the principal components are determined by mixtures of technical variables (e.g., media activity, voting effectiveness, age). No single feature fully dominates the spatial arrangement; both principal components represent blends of the anonymous and/or spatial features.  
Crucially, attributes such as promise fulfillment or group affiliation, which are easily interpreted in the avatar-based view, are not visible in the PCA scatter plot geometry unless they are explicitly included in the projection. It highlights the interpretability advantage of spatial avatars: only by explicit mapping can the user guarantee that key variables are visually represented.

\paragraph{Insights and sensitivity analysis:}
Changing the assignment of features (e.g., switching "fulfillment rate" from an avatar trait to a spatial coordinate or PCA) alters both the visual arrangement and interpretability of the visualization. Spatial avatars thus enable flexible, user-driven mapping, supporting either interpretability or structural discovery as needed.

\vspace{1em}

\subsection{Medium-scale illustration: Synthetic soft drinks dataset (N=100)}\label{sec:softdrinks}

The synthetic soft drinks dataset comprises 100 fictional beverages generated to simulate market diversity with three main product groups (classic, premium, and extreme). Each beverage is described by intuitive sensory attributes (sweetness, fizziness, overall rating, and color intensity) and technical features (price and caffeine). The \texttt{group\_numeric} variable encodes the product segment, used only for color coding; each sample is also labeled by name.

\begin{figure}[h!]
    \centering
    \includegraphics[width=.75\linewidth]{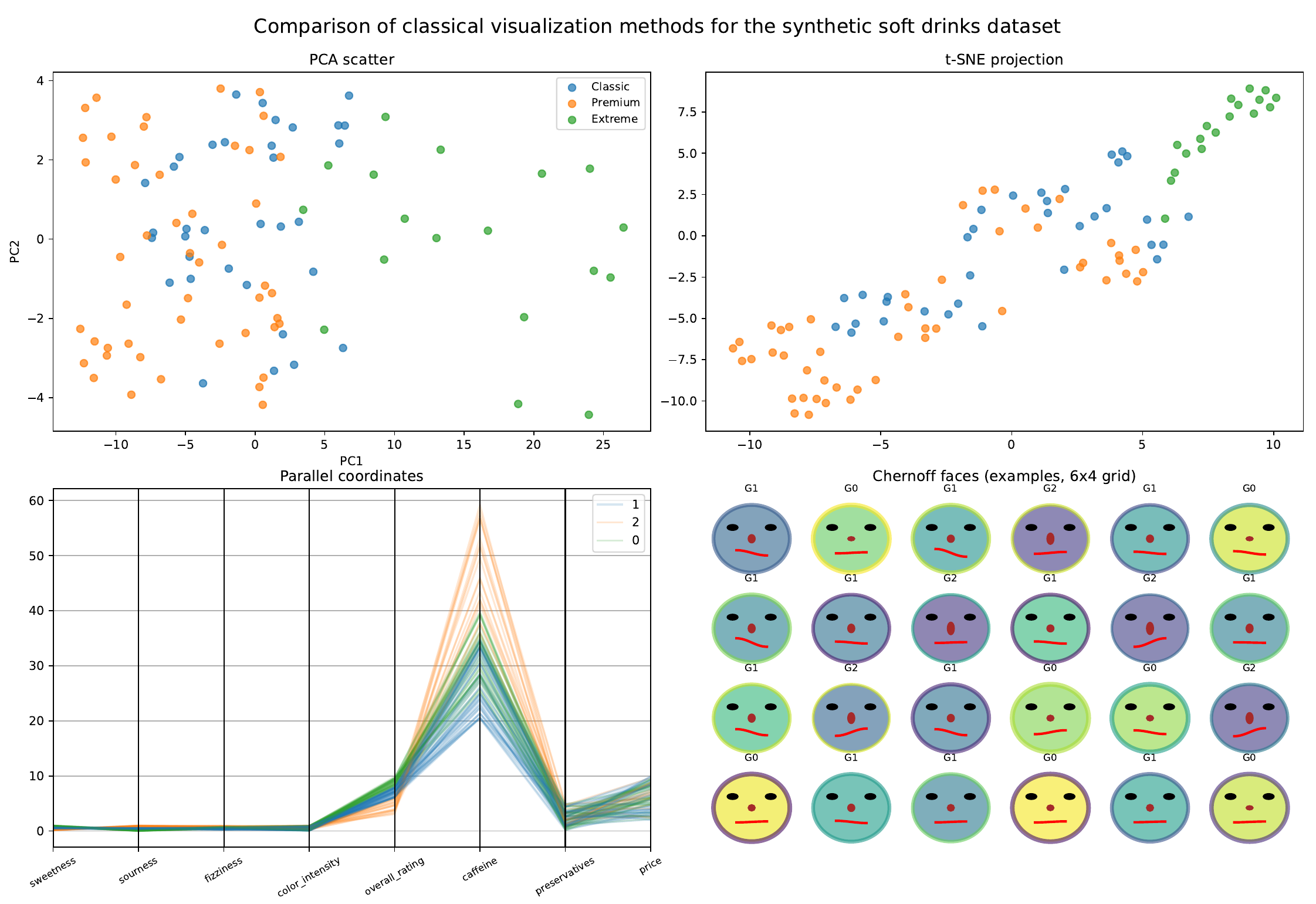}
    \caption{
        Comparison of classical visualization methods for the synthetic soft drinks dataset.
        \textbf{Top left:} PCA scatter plot;  
        \textbf{Top right:} t-SNE projection;  
        \textbf{Bottom left:} parallel coordinates plot;  
        \textbf{Bottom right:} Chernoff faces grid (6$\times$4) for a subset of beverages, colored by group.
        While PCA and t-SNE reveal some group structure, and parallel coordinates enable visualization of feature distributions, Chernoff faces offer a glimpse into sensory profiles of selected products. However, none of these methods provides a direct, interpretable synthesis of both product similarity and profile, as does the spatial avatar approach.
    }
    \label{fig:drinks_classical}
\end{figure}

To provide context for the effectiveness of our approach, Figure~\ref{fig:drinks_classical} shows the synthetic drinks dataset visualized using four classical techniques: PCA scatter plot, t-SNE projection, parallel coordinates, and a Chernoff faces grid for selected products. PCA and t-SNE projections highlight clear clustering by product type, with the principal components and t-SNE axes separating the main beverage groups. However, interpreting which specific features drive the separation or identifying outliers based on concrete sensory traits remains challenging. Parallel coordinates enable a detailed comparison of feature values but quickly become cluttered at this scale. Chernoff faces communicate some intuitive traits yet are difficult to interpret en masse and do not encode structural relationships between products.

\paragraph{Variable assignment:}
Intuitive features (sweetness, fizziness, overall rating, color intensity) were mapped to avatar visual traits (e.g., smile, nose length, skin color, hair color). Technical features (price, caffeine) were used for spatial coordinates (X and Y axes). The remaining attributes—sourness, sugar, citric acid, CO$_2$ pressure, preservatives, and color intensity—were included as anonymous variables for PCA-based analysis. This assignment is reflected in the figures below.

\begin{figure}[h!]
    \centering
    \includegraphics[width=.7\linewidth]{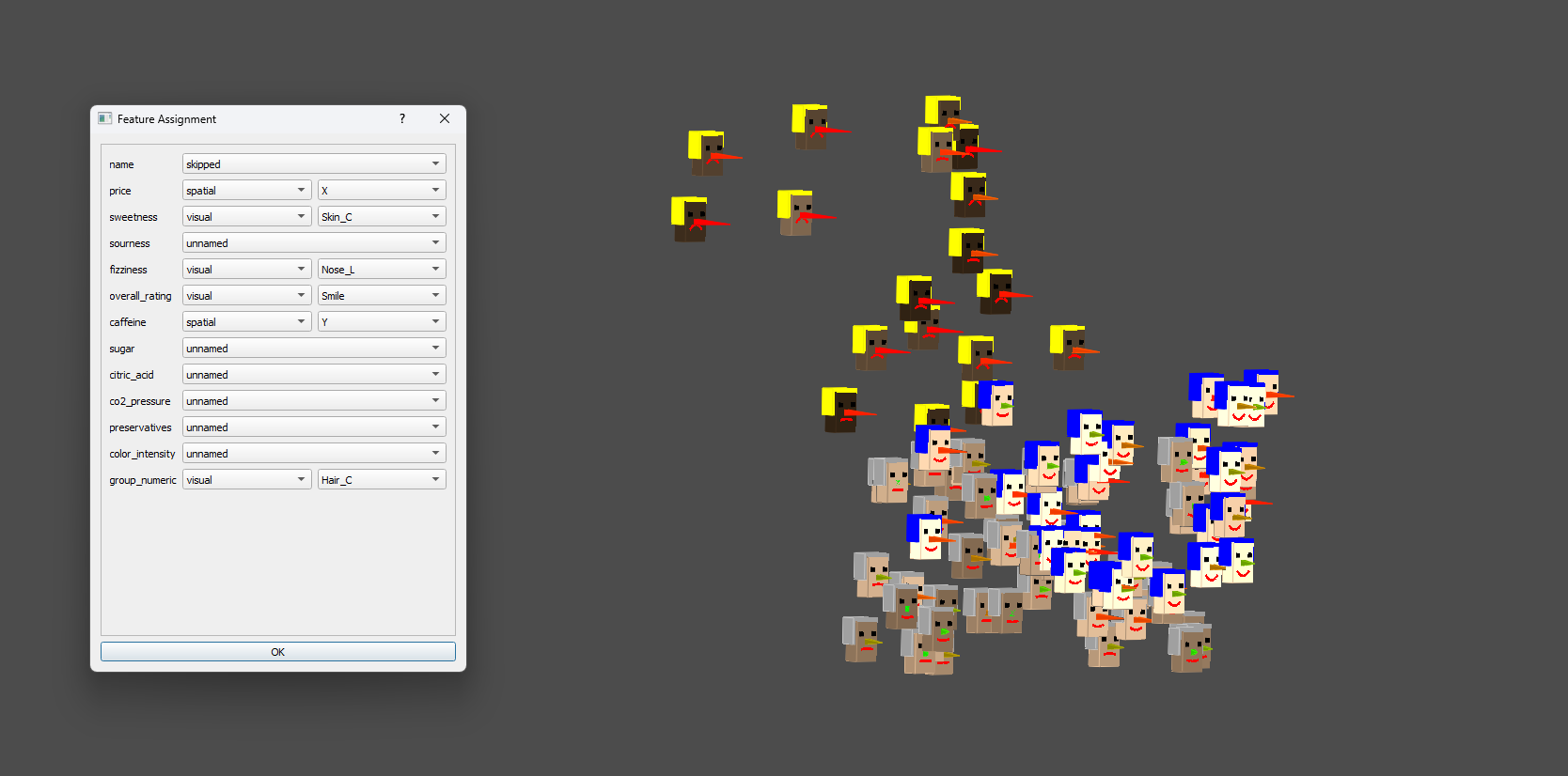}
    \caption{
        Spatial avatar visualization of the synthetic soft drinks dataset.
        Avatar features encode intuitive sensory attributes: sweetness (skin color), fizziness (nose length), overall rating (smile curvature), and product group (hair color).
        Spatial coordinates reflect technical properties: price (X) and caffeine (Y).
        This visualization enables immediate comparison of product types and detection of unusual profiles, with both group structure and individual differences readily interpretable.
    }
    \label{fig:softdrinks_swarm}
\end{figure}

\begin{figure}[h!]
    \centering
        \includegraphics[width=.75\linewidth]{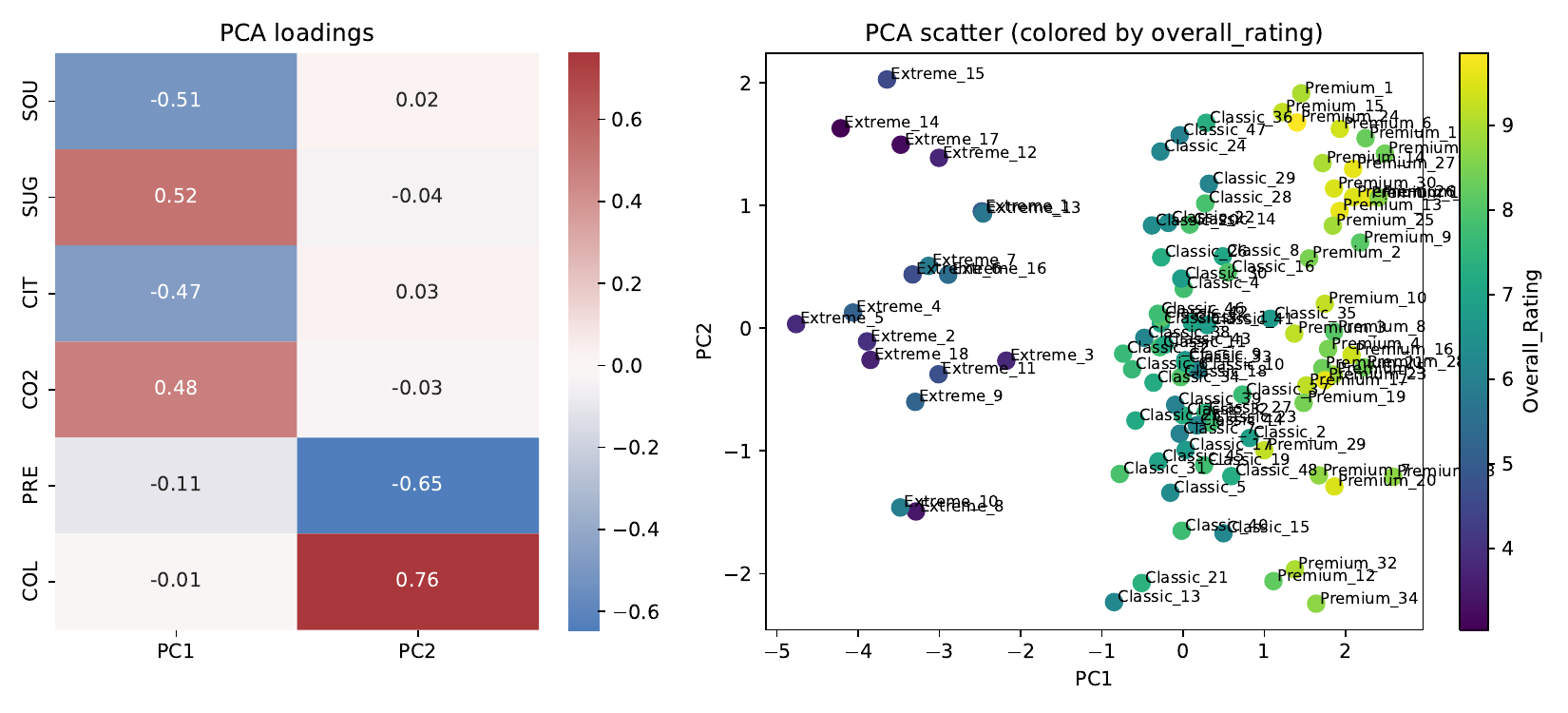}\\
        \includegraphics[width=.75\linewidth]{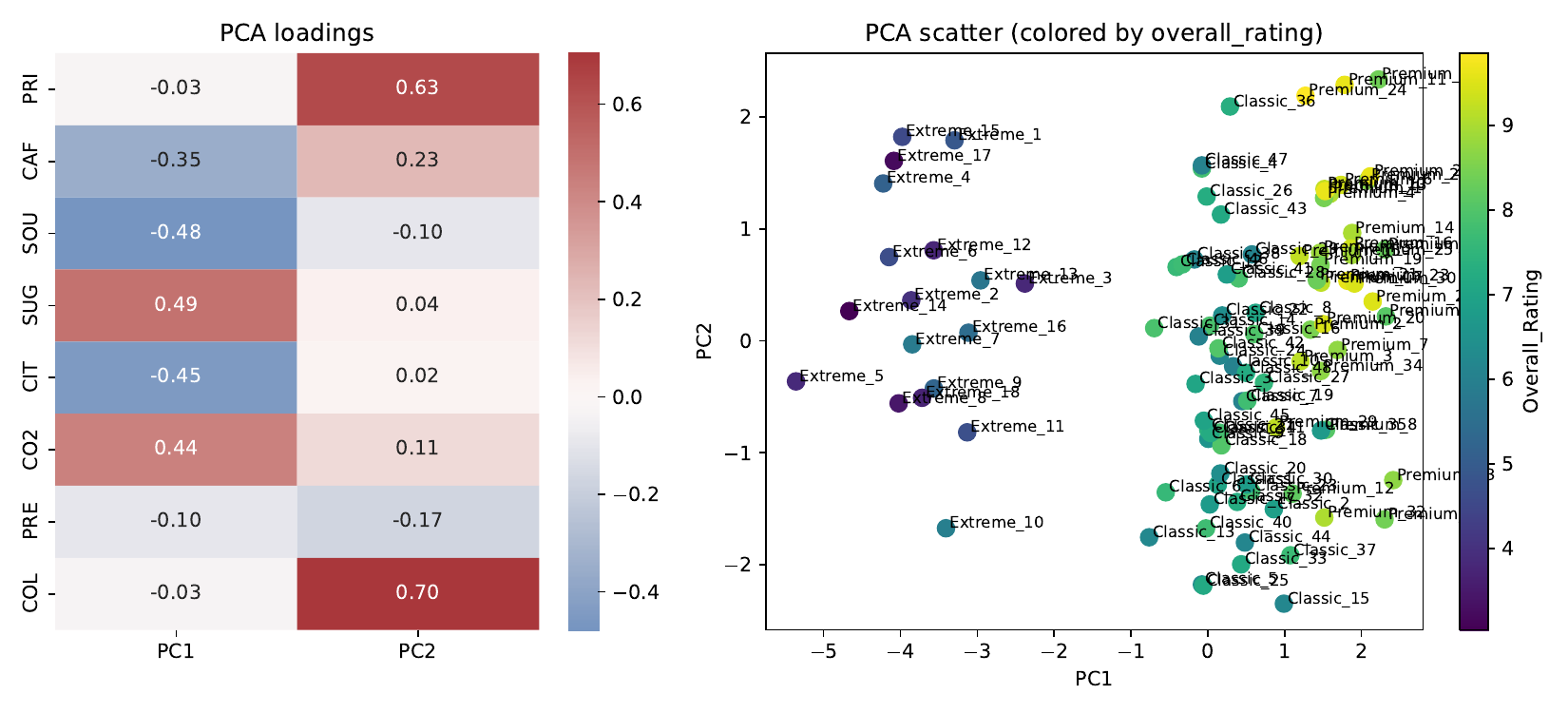}
    \caption{
        Effect of variable assignment on principal component analysis for the synthetic soft drinks dataset.
        \textbf{(top)} PCA on anonymous features only.
        \textbf{(bottom)} PCA on all features assigned to spatial coordinates or left anonymous (all except avatar features).
        In both cases, intuitive variables such as overall rating are not directly represented in the geometry unless they are included in the feature set, and their effect may only be visible through color.
        Variable abbreviations: SOU---sourness, SUG---sugar, CIT---citric acid, CO2---CO$_2$ pressure, PRE---preservatives, COL---color intensity, PRC---price, CAF---caffeine.
    }
    \label{fig:softdrinks_pca_analysis}
\end{figure}

\paragraph{PCA loadings and spatial interpretation:}
Principal component analysis was performed under two scenarios: (top) only on anonymous features (not mapped to avatars or spatial coordinates), and (bottom) on all features assigned to spatial coordinates or left anonymous. 
Figure~\ref{fig:softdrinks_pca_analysis} presents heatmaps of PCA loadings and corresponding scatter plots for both cases.

Notably, in both PCA scenarios, clear clusters emerge corresponding to beverage groups, as principal components are determined by mixtures of compositional variables (e.g., sourness, sugar, citric acid, preservatives). However, while structural groupings are evident, the interpretation of which sensory or technical attributes drive the group separation and direct identification of outliers remains less intuitive than in the spatial avatar visualization (Fig.~\ref{fig:softdrinks_swarm}), where both group membership and profile details are visible at a glance.

\paragraph{Insights and sensitivity analysis:}
This example demonstrates that, although dimensionality reduction techniques like PCA and t-SNE can recover the underlying group structure when relevant features dominate the variance, only the spatial avatar approach allows for the immediate interpretation of which specific sensory traits differentiate products, providing direct, profile-based insight. Switching a feature, such as ''sweetness'' or ''fizziness'', from an avatar trait to an anonymous (PCA) feature alters the PCA scatter plot and principal component structure but at the expense of immediate interpretability in the avatar view. This flexibility allows users to prioritize either structural discovery or interpretability as needed.

\vspace{1em}

\subsection{Large-scale real-world: Wine dataset (N=1599)}\label{sec:winedata}

We evaluated the method on the \textit{Vinho Verde} dataset~\cite{Cortez09,wine_quality_186}, which contains 1,599 samples of Portuguese red wines, each characterized by 12 physicochemical and sensory attributes measured in laboratory analysis. This dataset is widely used as a benchmark for multivariate modeling and classification tasks in oenology.

\begin{figure}[h!]
    \centering
    \includegraphics[width=.75\linewidth]{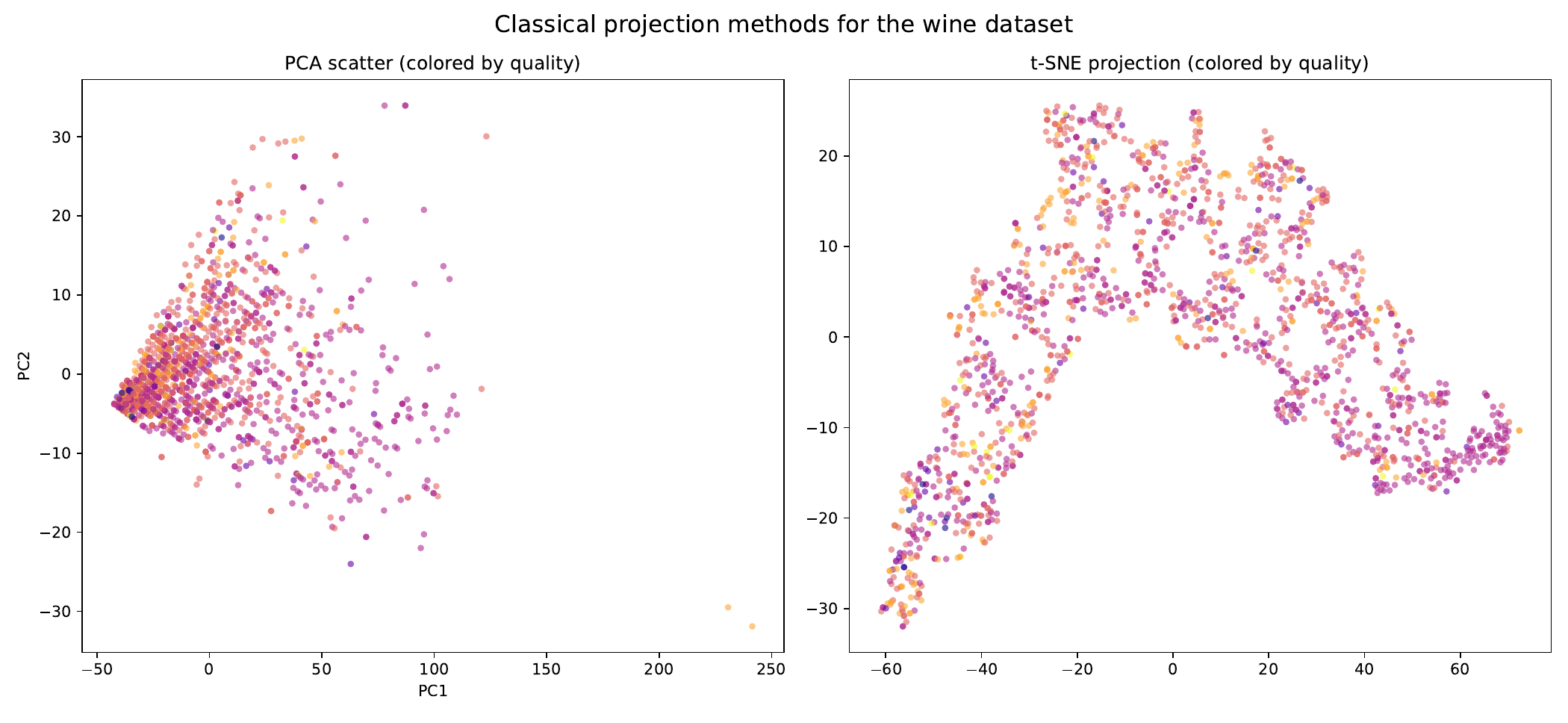}
    \caption{
        Classical projection methods for the wine dataset.
        \textbf{Left:} PCA scatter plot colored by quality rating;
        \textbf{Right:} t-SNE projection colored by quality.
        For large datasets such as this, other classical methods (e.g., parallel coordinates or glyph-based Chernoff faces) rapidly become uninterpretable due to severe overplotting and visual clutter and are therefore omitted.
    }
    \label{fig:wines_classical}
\end{figure}

Classical visualization techniques such as PCA and t-SNE scatter plots (Figure~\ref{fig:wines_classical}) provide a global view of data structure. However, for the wine dataset, no distinct clusters or quality-based groupings are apparent; samples of differing quality are dispersed throughout both projections. Other approaches, such as parallel coordinates or glyph-based Chernoff faces, are not shown, as they become uninformative for $N \gg 100$ due to overplotting and loss of interpretability.

\paragraph{Variable assignment:}
Quality, alcohol content, and residual sugar were assigned as avatar features (e.g., smile, face color, nose length), as these are familiar and interpretable for non-experts. Four technical variables (volatile acidity, citric acid, residual sugar, sulfates) were assigned to spatial coordinates (X, Y, Z, T). All remaining technical features were omitted from the spatial mapping in this scenario.

\begin{figure}[h!]
    \centering
    \includegraphics[width=.7\linewidth]{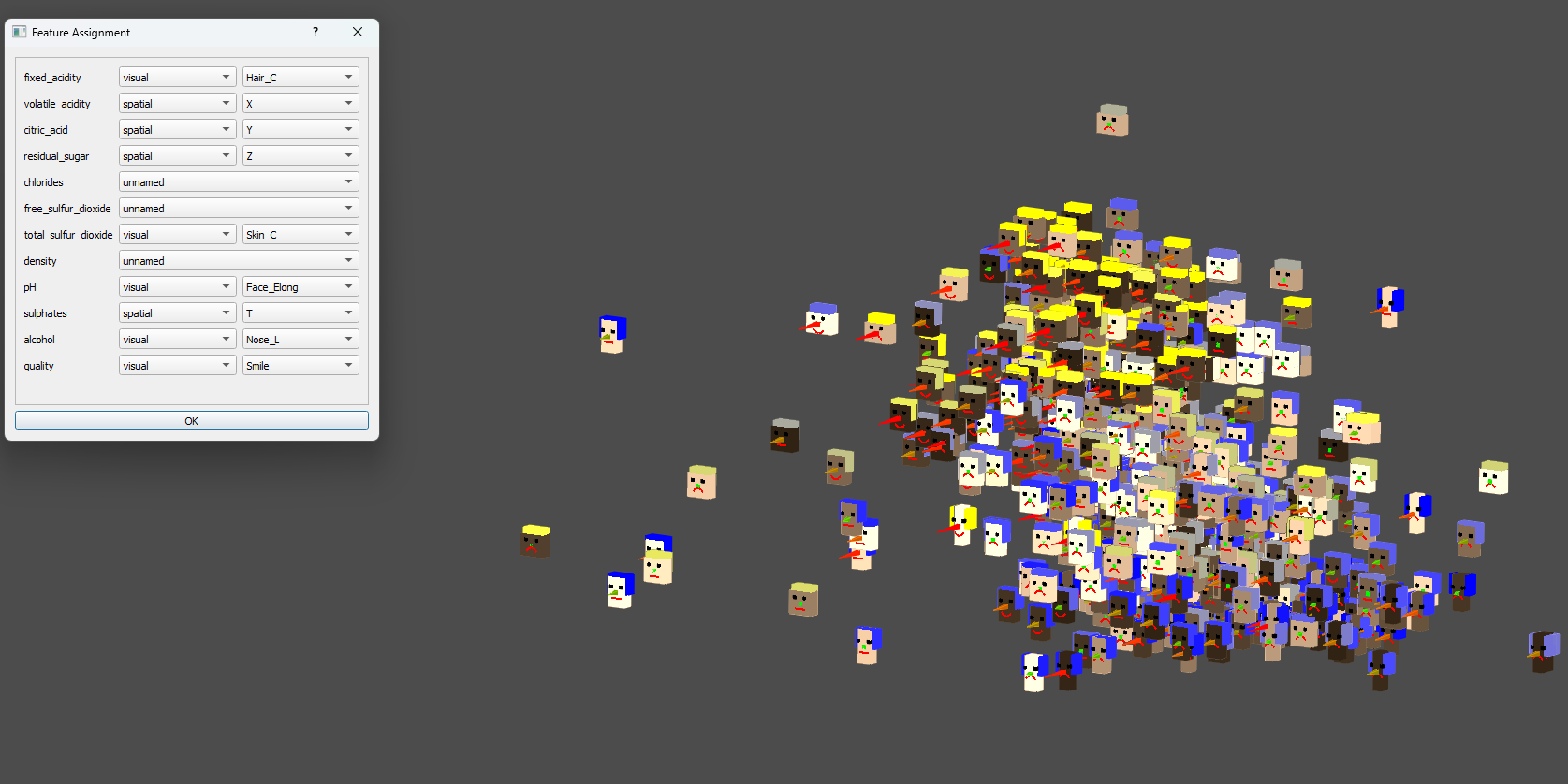}
    \caption{
        Spatial avatar visualization of the Vinho Verde wine dataset.
        Avatar features encode quality, alcohol content, and residual sugar; spatial coordinates reflect volatile acidity, citric acid, residual sugar, and sulfates.
        This mapping enables immediate identification of wines with unusual or extreme property profiles, as well as direct interpretation of both spatial position and visual traits.
    }
    \label{fig:wines_swarm}
\end{figure}

\begin{figure}[h!]
    \centering
        \includegraphics[width=.75\linewidth]{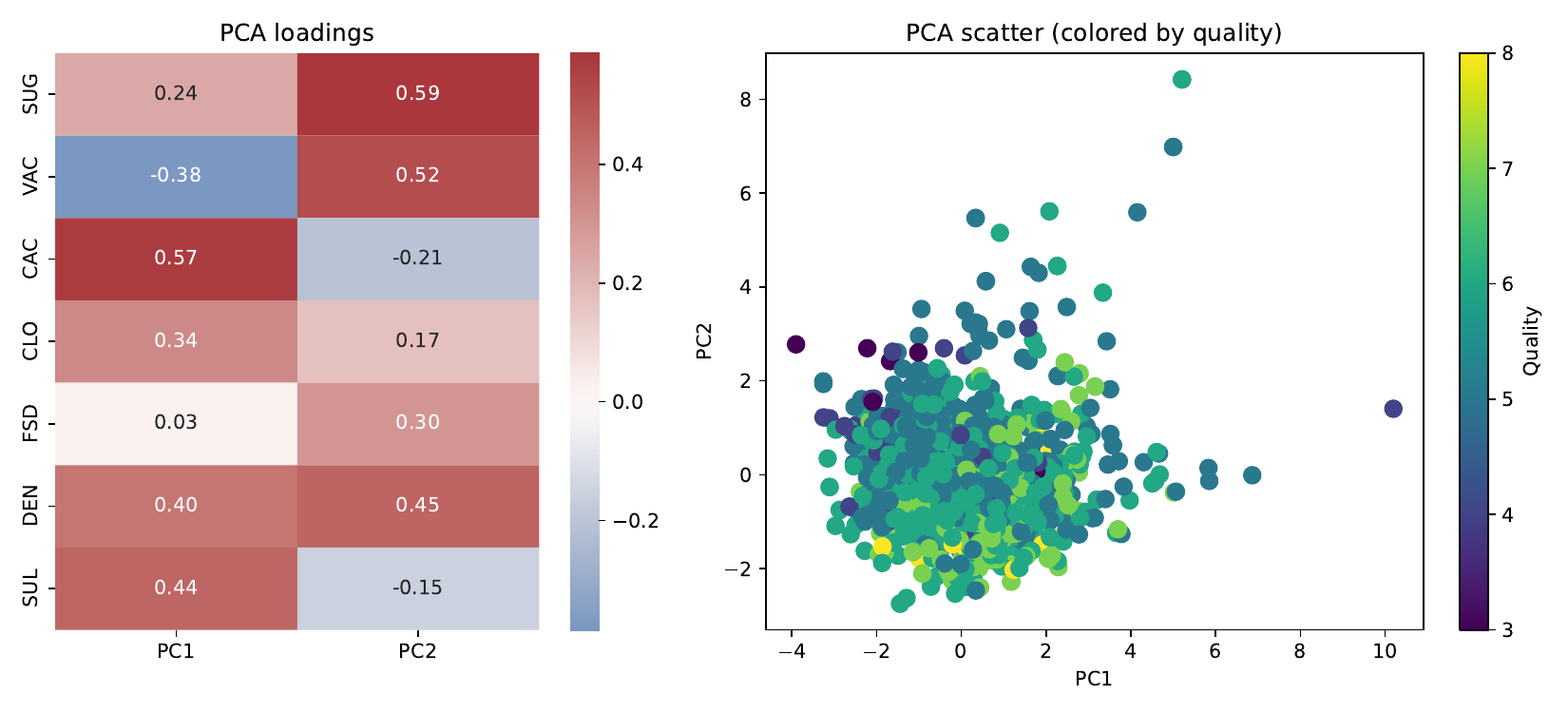}
    \caption{
        Principal component analysis for the wine dataset based on all technical variables except avatar-assigned ones.
        The first two principal components are mixtures of several physicochemical properties (PC1: volatile acidity 0.57, density 0.40, sulfates 0.44; PC2: residual sugar 0.59, citric acid 0.52, density 0.45).
        No clear clusters or quality-based groupings emerge---high- and low-quality wines are scattered across the projection.
        It highlights the limitation of PCA-based projections for interpretability compared to explicit, user-driven assignments in spatial avatars.
        Variable abbreviations: VAC---volatile acidity, CAC---citric acid, CLO---chlorides, FSD---free sulfur dioxide, SUL---sulfates, DEN---density, SUG---residual sugar.
    }
    \label{fig:wines_pca_analysis}
\end{figure}

\paragraph{PCA loadings and spatial interpretation:}
For comparison, principal component analysis was performed on all technical features except those assigned to avatar traits. Figure~\ref{fig:wines_pca_analysis} presents a heatmap of PCA loadings and the corresponding scatter plot, with samples colored by quality.

In the PCA projection, the principal components reflect combinations of technical features, and the spatial arrangement does not reveal distinct clusters or gradients by quality. By contrast, the spatial avatar visualization (Fig.~\ref{fig:wines_swarm}) ensures that critical consumer-facing attributes remain visible and interpretable due to explicit user assignment of axes.

\paragraph{Insights and sensitivity analysis:}
Assigning ''quality'' or ''alcohol'' as a spatial or PCA feature, rather than as an avatar trait, would affect the composition of principal components but reduce the immediate interpretability of these key attributes---demonstrating the practical benefit of flexible, user-driven assignment in the proposed approach.

\section{Discussion}\label{sec:discussion}

This study introduces a flexible and interpretable framework for multidimensional data visualization, placing the user at the center of the exploratory process. The core innovation lies in the semantic division of variables: Users can assign the most meaningful interpretable features (such as quality, popularity, or sensory attributes) to avatar characteristics, while more abstract or technical variables (e.g., chemical or behavioral parameters) are mapped to spatial coordinates. Unlike traditional approaches, such as PCA or parallel coordinates---which treat all variables equally and often obscure critical features---our method allows for user-driven, context-sensitive mapping that supports practical interpretability.

Spatial avatars further extend classical glyph-based visualization by embedding expressive, user-configurable avatars in a fully interactive 3D environment. Users can dynamically explore the data cloud, rotate and zoom into clusters, and rapidly test hypotheses by reassigning features to avatars or axes. This interactive capacity enables both the detection of structural patterns and the immediate comparison of individual profiles---capabilities not readily available in static projection-based visualizations.

Empirical results across synthetic and real-world datasets---including fictional politicians, synthetic soft drinks, and the Vinho Verde wine collection---demonstrate that spatial avatars make group structure, outliers, and archetypes directly observable and interpretable. In contrast, classical PCA-based projections often mix multiple technical variables into principal components, rendering key features invisible or ambiguous from the user's perspective.

By empowering users to control the mapping of features and explore data interactively, the spatial avatar framework bridges the gap between statistical rigor and practical, human-centered interpretation. The system, implemented as an open-source plugin for the \textit{dpVision} platform, is adaptable to various analytical goals, data types, and user preferences.

\subsection{Limitations}

While the proposed approach offers significant advantages, several limitations should be acknowledged. The semantic assignment of variables to ''intuitive'' or ''technical'' categories relies on user expertise and task context, introducing some subjectivity; different users or domains may lead to different choices and, thus, different interpretations. The number of avatar visual features is limited, restricting the range of intuitive variables that can be displayed directly, which may necessitate dimensionality reduction or the omission of some information. When many variables are treated as ''anonymous'' and incorporated via PCA, interpretability can be lost since principal components represent linear combinations rather than original features. Moreover, expressive avatars may introduce emotional or cognitive biases, and for very large or high-dimensional datasets, visual clutter or computational limits could arise. Finally, variable assignment is currently manual; integrating automated or user-guided assignment tools remains an important direction for further work.

Although a formal user study has not yet been conducted, the method has been informally evaluated among collaborating researchers. These preliminary trials indicated that certain avatar features---such as face color or hair color---are readily distinguished and effective for communicating differences, while others, including smile curvature or subtle facial traits, may be less easily perceived. This observation highlights the importance of meticulous glyph design and perceptual testing to enhance interpretability. Future work will involve systematic user studies to quantitatively assess the recognizability of individual avatar features and guide further refinements of the visual encoding.

Addressing these limitations is a natural path for development, including more robust assignment strategies, alternative glyphs to reduce bias, and enhanced support for complex or large-scale datasets.

\subsection{Advantages and utility of spatial avatars}

Compared to conventional dimensionality reduction methods, the spatial avatar approach provides direct, user-controlled interpretability by allowing the explicit mapping of intuitive features to avatar traits. It supports immediate visual recognition of unusual or exemplary cases and leverages innate human pattern recognition skills. Technical variables projected onto spatial coordinates reveal underlying structures, clusters, or outliers, while avatars simultaneously communicate key user-relevant attributes. Interactive mapping and dynamic exploration enable users to iteratively test hypotheses, adjust variable assignments, and visualize the impact on the data structure.

Empirical case studies (see Section~Results) show that spatial avatars reveal patterns and support interpretations that are less accessible or even hidden in standard visualization techniques such as PCA. Sensitivity analyses demonstrate how variable assignment impacts interpretability and cluster formation, underscoring the method’s flexibility and value across diverse analytical contexts.

\subsection{Future work}\label{sec:future_work}

Several directions for further development are envisaged. One approach is to expand the repertoire of avatar types: while Chernoff-like faces take advantage of human abilities to perceive facial traits, alternative glyphs---such as neutral geometric shapes or biologically inspired forms---could reduce emotional bias or adapt to new domains. Another area is the automation of variable assignment; future versions could integrate heuristics or machine learning to suggest optimal mapping based on data properties or user priorities. Scalability remains an open challenge: dynamic filtering, clustering, sampling, or zoomable detail views may help maintain interpretability in larger data collections, possibly in combination with statistical overviews. Ultimately, formal usability studies are necessary to validate the approach in controlled settings and to inform refinements for diverse application areas.

\section{Conclusion}\label{sec:conclusion}

In this paper, we presented a novel method for visualizing high-dimensional data by integrating two natural strengths of human perception: spatial understanding and intuitive recognition of visual patterns. Our approach distinguishes between intuitive and abstract data dimensions, mapping them in different ways for optimal interpretability. Intuitive features are assigned to avatar characteristics, while technical features are reduced and projected into 3D space.

Implemented as a plugin for the open-source \textit{dpVision} platform, the method was demonstrated on synthetic and real-world datasets. The combination of avatar-based glyphs and interactive 3D navigation provides an intuitive and flexible way to explore complex data structures, facilitating the discovery of patterns and individualized interpretation.

Future work will focus on formal usability evaluation, the development of alternative glyph styles, automated variable mapping, and further improvements to scalability. By enabling users to tailor visual representation to their needs, our framework advances a human-centered paradigm for the visual analysis of complex multidimensional data.

\section*{Author Contributions}
Conceptualization, L.L.; methodology, L.L.; software, L.L. and D.P.; validation, L.L. and D.P.; formal analysis, L.L.; investigation, D.P.; resources, L.L.; data curation, D.P.; writing---original draft preparation, L.L. and D.P.; writing---review and editing, L.L. and D.P.; visualization, D.P.; supervision, L.L.; project administration, D.P. All authors have read and agreed to the published version of the manuscript.

\section*{Funding}
This research received no external funding.

\section*{Data Availability Statement}

The source code for the plugin is available at \url{https://github.com/iitis/N-Dim-view}. All synthetic datasets used in this study, together with the additional code required to reproduce the experiments and visualizations, are included in this repository. The real-world wine dataset used is publicly available and formally cited in the References. The source code for the dpVision framework is available at \url{https://github.com/pojdulos/dpVision}.

\section*{Conflicts of Interest}
The authors declare no conflict of interest.

\section*{Acknowledgments}
The authors would like to thank Krzysztof Domino, PhD, for his valuable advice and motivation, which supported the creation of this article.

\section*{List of Abbreviations}
\begin{tabular}{ll}
PCA   & Principal Component Analysis \\
PC    & Principal Component \\
t-SNE & t-distributed Stochastic Neighbor Embedding \\
dpVision & Open-source Data Visualization Platform \\
GUI   & Graphical User Interface \\
SO(2) & Special Orthogonal Group in 2D (if used in text) \\
SO(4) & Special Orthogonal Group in 4D (if used in text) \\
SVD   & Singular Value Decomposition (if used in text) \\
\end{tabular}


\begin{thebibliography}{10}
\providecommand{\url}[1]{#1}
\csname url@samestyle\endcsname
\providecommand{\newblock}{\relax}
\providecommand{\bibinfo}[2]{#2}
\providecommand{\BIBentrySTDinterwordspacing}{\spaceskip=0pt\relax}
\providecommand{\BIBentryALTinterwordstretchfactor}{4}
\providecommand{\BIBentryALTinterwordspacing}{\spaceskip=\fontdimen2\font plus
\BIBentryALTinterwordstretchfactor\fontdimen3\font minus \fontdimen4\font\relax}
\providecommand{\BIBforeignlanguage}[2]{{%
\expandafter\ifx\csname l@#1\endcsname\relax
\typeout{** WARNING: IEEEtran.bst: No hyphenation pattern has been}%
\typeout{** loaded for the language `#1'. Using the pattern for}%
\typeout{** the default language instead.}%
\else
\language=\csname l@#1\endcsname
\fi
#2}}
\providecommand{\BIBdecl}{\relax}
\BIBdecl

\bibitem{Chernoff1973}
\BIBentryALTinterwordspacing
H.~Chernoff, ``{The Use of Faces to Represent Points in K-Dimensional Space Graphically},'' \emph{Journal of the American Statistical Association}, vol.~68, no. 342, pp. 361--368, 1973. [Online]. Available: \url{http://dx.doi.org/10.2307/2284077}
\BIBentrySTDinterwordspacing

\bibitem{Lardelli}
\BIBentryALTinterwordspacing
M.~Lardelli, ``Beyond chernoff faces: Multivariate visualization with metaphoric 3d glyphs.'' [Online]. Available: \url{https://lardel.li/2017/07/beyond-chernoff-faces-multivariate-visualization-with-metaphoric-3d-glyphs.html}
\BIBentrySTDinterwordspacing

\bibitem{POJDA2025102093}
\BIBentryALTinterwordspacing
D.~Pojda, M.~Żarski, A.~A. Tomaka, and L.~Luchowski, ``dp{V}ision: {E}nvironment for multimodal images,'' \emph{SoftwareX}, vol.~30, p. 102093, 2025. [Online]. Available: \url{http://dx.doi.org/10.1016/j.softx.2025.102093}
\BIBentrySTDinterwordspacing

\bibitem{dpvision}
\BIBentryALTinterwordspacing
D.~Pojda, ``dp{V}ision (data processing for vision),'' opensource software repository, accessed: 2025-03-02. [Online]. Available: \url{https://doi.org/10.5281/zenodo.13944334}
\BIBentrySTDinterwordspacing

\bibitem{Cortez09}
P.~Cortez, J.~Teixeira, A.~Cerdeira, F.~Almeida, T.~Matos, and J.~Reis, ``{Using Data Mining for Wine Quality Assessment},'' in \emph{Proceedings of the International Conference on Discovery Science}, Berlin-Heidelberg, Germany, 2009, pp. 66--79.

\bibitem{wine_quality_186}
P.~Cortez, A.~Cerdeira, F.~Almeida, T.~Matos, and J.~Reis, ``{Wine Quality},'' UCI Machine Learning Repository, 2009.

\bibitem{Jolliffe2011}
I.~Jolliffe, \emph{Principal Component Analysis}.\hskip 1em plus 0.5em minus 0.4em\relax Berlin, Heidelberg: Springer Berlin Heidelberg, 2011, pp. 1094--1096.

\bibitem{Maaten2008}
\BIBentryALTinterwordspacing
L.~van~der Maaten and G.~Hinton, ``Visualizing data using t-sne,'' \emph{Journal of Machine Learning Research}, vol.~9, no.~86, pp. 2579--2605, 2008. [Online]. Available: \url{http://jmlr.org/papers/v9/vandermaaten08a.html}
\BIBentrySTDinterwordspacing

\bibitem{McInnes2018}
L.~McInnes, J.~Healy, N.~Saul, and L.~Großberger, ``Umap: Uniform manifold approximation and projection,'' \emph{Journal of Open Source Software}, vol.~3, no.~29, p. 861, 2018.

\bibitem{d_ocagne}
\BIBentryALTinterwordspacing
M.~d'Ocagne, ``Les coordonnées parallèles de points,'' \emph{Nouvelles annales de mathématiques : journal des candidats aux écoles polytechnique et normale}, vol.~6, pp. 493--502, 1887. [Online]. Available: \url{www.numdam.org/item/NAM_1887_3_6__493_0/}
\BIBentrySTDinterwordspacing

\bibitem{Inselberg1985}
A.~Inselberg, ``The plane with parallel coordinates,'' \emph{The Visual Computer}, vol.~1, no.~2, pp. 69--91, 1985.

\bibitem{Andrews}
D.~F. Andrews, ``Plots of high-dimensional data,'' \emph{Biometrics}, 1972.

\bibitem{Cleveland1993}
W.~S. Cleveland, \emph{Visualizing Data}.\hskip 1em plus 0.5em minus 0.4em\relax Hobart Press, 1993.

\bibitem{Cook05}
A.~Buja, D.~Cook, D.~Asimov, and C.~Hurley, ``14 - computational methods for high-dimensional rotations in data visualization,'' in \emph{Data Mining and Data Visualization}, ser. Handbook of Statistics, C.~Rao, E.~Wegman, and J.~Solka, Eds.\hskip 1em plus 0.5em minus 0.4em\relax Elsevier, 2005, vol.~24, pp. 391--413.

\bibitem{Cook2020SliceTour}
U.~Laa, D.~Cook, and G.~V. and, ``A slice tour for finding hollowness in high-dimensional data,'' \emph{Journal of Computational and Graphical Statistics}, vol.~29, no.~3, pp. 681--687, 2020.

\bibitem{PFLUGHOEFT2024103911}
K.~A. Pflughoeft, F.~M. Zahedi, and Y.~Chen, ``Data avatars: A theory-guided design and assessment for multidimensional data visualization,'' \emph{Information \& Management}, vol.~61, no.~2, p. 103911, 2024.

\end{thebibliography}


\end{document}